\documentclass[runningheads]{llncs}

\usepackage{eccv}

\usepackage{eccvabbrv}

\usepackage{graphicx}
\usepackage{booktabs}

\usepackage[accsupp]{axessibility}  

\usepackage{hyperref}

\usepackage{orcidlink}

\usepackage{colortbl}
\usepackage{amsmath}
\usepackage{amssymb}
\usepackage{booktabs}
\usepackage{enumitem}
\newcommand\crule[3][black]{\textcolor{#1}{\rule{#2}{#3}}}
\usepackage{booktabs, multirow} 
\definecolor{roadcolor}{RGB}{234,51,246}
\definecolor{sidewalkcolor}{RGB}{68,8,72}
\definecolor{parkingcolor}{RGB}{241,156,249}
\definecolor{othergroundcolor}{RGB}{160,32,76}
\definecolor{buildingcolor}{RGB}{246,202,69}
\definecolor{carcolor}{RGB}{111,149,238}
\definecolor{truckcolor}{RGB}{74,32,172}
\definecolor{bicyclecolor}{RGB}{136,227,242}
\definecolor{motorcyclecolor}{RGB}{37,59,146}
\definecolor{othervehiclecolor}{RGB}{96,81,242}
\definecolor{vegetationcolor}{RGB}{79, 173, 50}
\definecolor{trunkcolor}{RGB}{126, 65, 22}
\definecolor{terraincolor}{RGB}{171, 238, 105}
\definecolor{personcolor}{RGB}{234, 60, 49}
\definecolor{bicyclistcolor}{RGB}{234, 66, 195}
\definecolor{motorcyclistcolor}{RGB}{138, 42, 90}
\definecolor{fencecolor}{RGB}{238, 128, 69}
\definecolor{polecolor}{RGB}{252, 241, 161}
\definecolor{trafficsigncolor}{RGB}{233, 51, 35}
\definecolor{color1}{RGB}{176, 36, 24}
\definecolor{color2}{RGB}{119,185,0}
\definecolor{color3}{RGB}{0, 0, 200}
\definecolor{colorofteaser}{RGB}{176, 36, 24}

\definecolor{LightGrey}{rgb}{.9,.9,.9}
\definecolor{White}{rgb}{1.,0.,1.}
\definecolor{first}{rgb}{.8,.0,.0}
\definecolor{second}{rgb}{.0,.6,.0}
\definecolor{third}{rgb}{.0,.0,.8}
\definecolor{ceiling}{RGB}{214,  38, 40}   %
\definecolor{floor}{RGB}{43, 160, 4}     %
\definecolor{wall}{RGB}{158, 216, 229}  %
\definecolor{window}{RGB}{114, 158, 206}  %
\definecolor{chair}{RGB}{204, 204, 91}   %
\definecolor{bed}{RGB}{255, 186, 119}  %
\definecolor{sofa}{RGB}{147, 102, 188}  %
\definecolor{table}{RGB}{30, 119, 181}   %
\definecolor{tvs}{RGB}{160, 188, 33}   %
\definecolor{furniture}{RGB}{255, 127, 12}  %
\definecolor{objects}{RGB}{196, 175, 214} %
\definecolor{car}{rgb}{0.39215686, 0.58823529, 0.96078431}
\definecolor{bicycle}{rgb}{0.39215686, 0.90196078, 0.96078431}
\definecolor{motorcycle}{rgb}{0.11764706, 0.23529412, 0.58823529}
\definecolor{truck}{rgb}{0.31372549, 0.11764706, 0.70588235}
\definecolor{other-vehicle}{rgb}{0.39215686, 0.31372549, 0.98039216}
\definecolor{person}{rgb}{1.        , 0.11764706, 0.11764706}
\definecolor{bicyclist}{rgb}{1.        , 0.15686275, 0.78431373}
\definecolor{motorcyclist}{rgb}{0.58823529, 0.11764706, 0.35294118}
\definecolor{road}{rgb}{1.        , 0.        , 1.        }
\definecolor{parking}{rgb}{1.        , 0.58823529, 1.        }
\definecolor{sidewalk}{rgb}{0.29411765, 0.        , 0.29411765}
\definecolor{other-ground}{rgb}{0.68627451, 0.        , 0.29411765}
\definecolor{building}{rgb}{1.        , 0.78431373, 0.        }
\definecolor{fence}{rgb}{1.        , 0.47058824, 0.19607843}
\definecolor{vegetation}{rgb}{0.        , 0.68627451, 0.        }
\definecolor{trunk}{rgb}{0.52941176, 0.23529412, 0.        }
\definecolor{terrain}{rgb}{0.58823529, 0.94117647, 0.31372549}
\definecolor{pole}{rgb}{1.        , 0.94117647, 0.58823529}
\definecolor{traffic-sign}{rgb}{1.        , 0.        , 0.    }

\definecolor{barrier1}{RGB}{112,128,144}
\definecolor{bicycle1}{RGB}{220,20,60}
\definecolor{bus1}{RGB}{255, 127, 80}
\definecolor{car1}{RGB}{255, 158, 0}
\definecolor{const. veh.1}{RGB}{233, 150, 70}
\definecolor{motorcycle1}{RGB}{255,61,99}
\definecolor{pedestrian1}{RGB}{0,0,230}
\definecolor{traffic cone1}{RGB}{47,79,79}
\definecolor{trailer1}{RGB}{255,140,0}
\definecolor{truck1}{RGB}{255,99,71}
\definecolor{drive. suf.1}{RGB}{0,207,191}
\definecolor{other flat1}{RGB}{175,0,75}
\definecolor{sidewalk1}{RGB}{75,0,75}
\definecolor{terrain1}{RGB}{112,180,60}
\definecolor{manmade1}{RGB}{222,184,135}
\definecolor{vegetation1}{RGB}{0,175,0}

\begin{document}

\title{Hierarchical Temporal Context Learning for Camera-based Semantic Scene Completion}

\titlerunning{Hierarchical Temporal Context Learning for Camera-based SSC}

\author{Bohan Li\inst{1,2}  \and
Jiajun Deng\inst{3}  \and
Wenyao Zhang \inst{1,2}  \and \\
Zhujin Liang \inst{4}  \and
Dalong Du \inst{4}  \and
Xin Jin\inst{2}\thanks{Corresponding author} \and
Wenjun Zeng\inst{2}
}

\authorrunning{B. Li et al.}

\institute{Shanghai Jiao Tong University, Shanghai, China \and
Ningbo Institute of Digital Twin, Eastern Institute of Technology, Ningbo, China  \and The University of Adelaide, Adelaide, Australia \and PhiGent Robotics, Beijing, China
\\
\email{\{bohan\_li, wy\_zhang\}@sjtu.edu.cn, jiajun.deng@adelaide.edu.au, \\
\{zhujin.liang, dalong.du\}@phigent.ai, \\ \{jinxin, wenjunzengvp\}@eitech.edu.cn} }

\maketitle

\begin{figure}
\begin{center} 
\vspace{-15pt}
\includegraphics[width=0.99\linewidth]{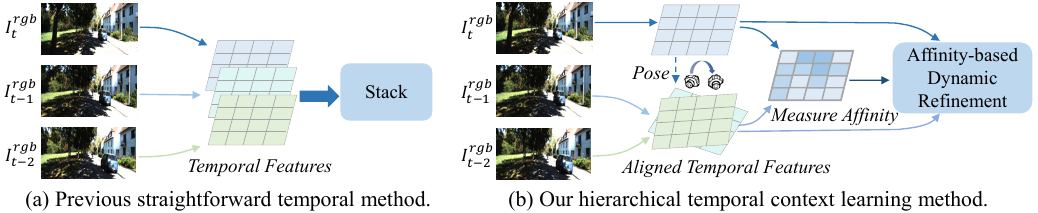}
 \vspace{-0pt}
\caption{Our hierarchical temporal context learning method versus previous straightforward temporal method (VoxFormer-T~\cite{li2023voxformer}) in semantic scene completion. 
}
\label{teaser1}
 \vspace{-30pt}
\end{center}
\end{figure}

\begin{abstract}
\vspace{-0pt}
Camera-based 3D semantic scene completion (SSC) is pivotal for predicting complicated 3D layouts with limited 2D image observations. The existing mainstream solutions generally leverage temporal information by roughly stacking history frames to supplement the current frame, such straightforward temporal modeling inevitably diminishes valid clues and increases learning difficulty. To address this problem, we present \textbf{HTCL}, a novel \textbf{H}ierarchical \textbf{T}emporal \textbf{C}ontext \textbf{L}earning paradigm for improving camera-based semantic scene completion.
The primary innovation of this work involves decomposing temporal context learning into two hierarchical steps: (a) cross-frame affinity measurement and (b) affinity-based dynamic refinement. Firstly, to separate critical relevant context from redundant information, we introduce the pattern affinity with scale-aware isolation and multiple independent learners for fine-grained contextual correspondence modeling. Subsequently, to dynamically compensate for incomplete observations, we adaptively refine the feature sampling locations based on initially identified locations with high affinity and their neighboring relevant regions. Our method ranks $1^{st}$ on the SemanticKITTI benchmark and even surpasses LiDAR-based methods in terms of mIoU on the OpenOccupancy benchmark. Our code is available on \url{https://github.com/Arlo0o/HTCL}.

  
\end{abstract}

\begin{figure}[!ht]
\vspace{-0pt}
\begin{center} 
	\begin{center}
		\includegraphics[width=0.99\linewidth]{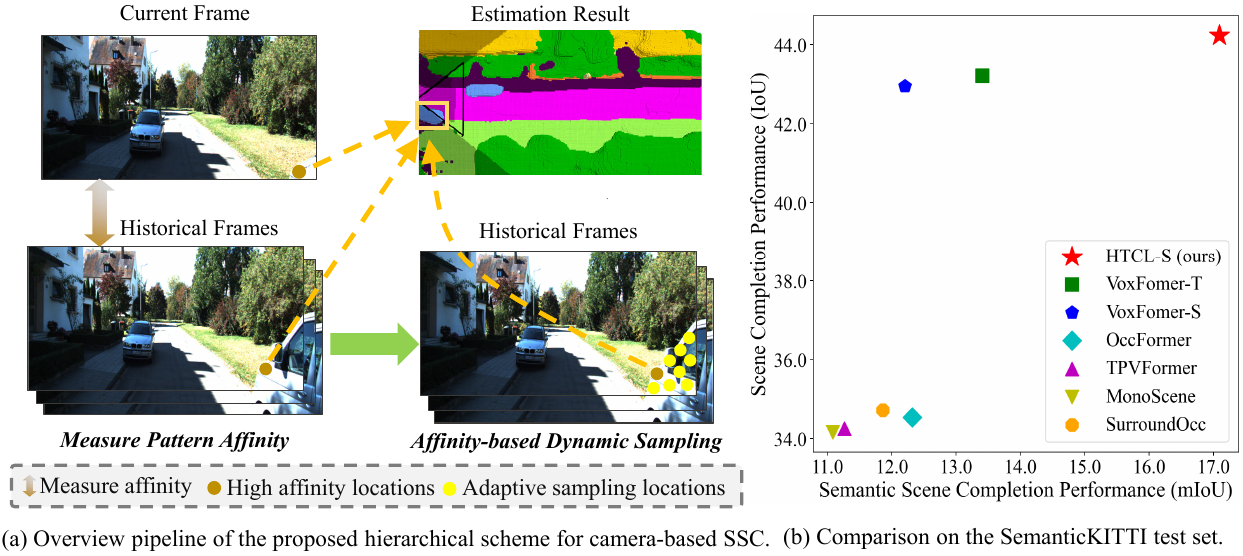}   
  
		\begin{tabular}{cccccc}	
			\multicolumn{6}{c}{
				\scriptsize
				\textcolor{bicycle}{$\blacksquare$}bicycle \textcolor{car}{$\blacksquare$}car \textcolor{motorcycle}{$\blacksquare$}motorcycle \textcolor{truck}{$\blacksquare$}truck \textcolor{other-vehicle}{$\blacksquare$}other-veh. \textcolor{person}{$\blacksquare$}person \textcolor{bicyclist}{$\blacksquare$}bicyclist \textcolor{motorcyclist}{$\blacksquare$}motorcyclist \textcolor{road}{$\blacksquare$}road
}
    \\
			\multicolumn{6}{c}{
				\scriptsize
                    \textcolor{parking}{$\blacksquare$}parking \textcolor{sidewalk}{$\blacksquare$}sidewalk \textcolor{other-ground}{$\blacksquare$}other.grd \textcolor{building}{$\blacksquare$}building \textcolor{fence}{$\blacksquare$}fence \textcolor{vegetation}{$\blacksquare$}vegetation \textcolor{trunk}{$\blacksquare$}trunk \textcolor{terrain}{$\blacksquare$}terrain \textcolor{pole}{$\blacksquare$}pole \textcolor{traffic-sign}{$\blacksquare$}traf.sign			
			}
		\end{tabular}	
 	\end{center}
        \vspace{-5pt}
	\captionof{figure}{(a) Overview pipeline of the proposed method, which measures contextual pattern affinity across temporal frames and dynamically samples relevant context. Our method shows promising performance in comprehending and completing semantic scenes even outside the camera's field of view, as indicated by the car highlighted with the yellow box.
 (b) Comparison with state-of-the-art camera-based semantic scene completion methods~\cite{li2023voxformer,zhang2023occformer,huang2023tri,cao2022monoscene,wei2023surroundocc} on the SemanticKITTI test set.
} 
\vspace{-30pt}
\label{teaser0}
\end{center}
\end{figure}

\section{Introduction}

The comprehension of holistic 3D scenes holds paramount importance in autonomous driving systems~\cite{li2023voxformer,huang2023tri,wei2023surroundocc}. 
This capability directly impacts the planning and obstacle avoidance functionalities of autonomous vehicles, thereby influencing their overall safety and efficiency.
However, due to the limitations of real-world sensors such as restricted field of view and measurement noise, this task remains a challenging problem. 
3D semantic scene completion (SSC) has been proposed to address the challenges by jointly inferring the geometry and semantics of the scenario from incomplete observations~\cite{zhang2018efficient,rist2021semantic,garbade2019two,roldao2020lmscnet,wu2020scfusion,cao2022monoscene,li2023voxformer,huang2023tri}.

Given the inherent 3D nature, numerous semantic scene completion (SSC) solutions~\cite{zhang2018efficient,rist2021semantic,garbade2019two,roldao2020lmscnet,wu2020scfusion} rely on LiDAR as the primary technique for precise 3D location measurement. Although the LiDAR sensor provides accurate depth information, its deployment introduces a large overhead for both the cost and manual efforts.
Therefore, it is necessary to explore an efficient approach for high-fidelity SSC with cost-effective devices.
This motivation led to the investigation of camera-based solutions, which are characterized by superior deployment efficiency and the abundance of rich visual context~\cite{cao2022monoscene,wei2023surroundocc,zhang2023occformer,huang2023tri}.

The early attempts in camera-based SSC methods~\cite{song2017semantic,cao2022monoscene,roldao20223d} commonly rely solely on the current frame, which can only provide very limited observation to recover the 3D geometry and semantics. 
To enrich contextual information, the pioneering work
VoxFormer-T~\cite{li2023voxformer} proposes to make use of temporal coherence by
stacking multiple history frames as supplements to the current frame. 
Nevertheless, as depicted in Figure~\ref{teaser1}, this approach assumes that temporal features from different viewpoints are originally corresponded at the pixel level, thereby just using a straightforward aggregation.
However, the shared semantic content undergoes uncertain positional changes across different perspectives. Therefore, directly integrating images from different timestamps may result in blurred predictive information. This ambiguity compromises the stability of semantic occupancy predictions and imposes difficulties on temporal modeling.

To alleviate this issue, we propose a new Hierarchical Temporal Context Learning (HTCL) paradigm, which improves the temporal feature aggregation for accurate 3D semantic scene completion. 
HTCL takes RGB images from different timestamps to hierarchically infer the 3D semantic occupancy for fine-grained scene comprehension. As depicted in Figure~\ref{teaser0} (a), our hierarchical temporal context modeling includes two sequential steps: (1) we explicitly measure the contextual pattern affinity between the current and historical frames, highlighting the most relevant patterns; (2) we adaptively refine the sampling locations based on the preliminary high-affinity locations and their nearby relevant context to dynamically compensate for incomplete observations.
Specifically, we first leverage epipolar homo-warping to explicitly align the temporal invariant feature representations and establish temporal feature volumes to fully maintain fine-grained context.
Then, we introduce \textit{scale-aware isolation} and \textit{incorporation of diverse independent learning} in the cross-frame affinity measurement to facilitate better affinity distribution modeling in SSC.
Subsequently, to dynamically compensate for incomplete observations, we resort to investigating the critical locations with high affinities and the neighboring relevant context. Technically, a multi-level deformable 3D block conditioned with the affinity weights is involved to adaptively refine the sampling locations.
Finally, a weighted voxel cross-attention is introduced to aggregate the reliable temporal content.

We conduct extensive experiments to validate the merits of our proposed HTCL. As shown in Figure~\ref{teaser0}(b), our proposed method achieves significant superiority over existing camera-based methods in terms of geometry (IoU) and semantics (mIoU). Remarkably, our camera-based approach outperforms state-of-the-art VoxFormer-T~\cite{li2023voxformer} 
on the SemanticKITTI benchmark and even surpasses LiDAR-based methods on the OpenOccuapcny benchmark in terms of mIoU. 
Our main contributions are summarized as follows:

\begin{itemize}

\item A temporal context learning paradigm with a hierarchical scheme to fully exploit dynamic and dependable 3D semantic scene completion.

\item An affinity measurement strategy with scale-aware isolation and multiple independent learners for fine-grained contextual correspondence modeling. 

\item An affinity-based dynamic refinement schedule to reassemble the temporal content and adaptively compensate for incomplete observations.

\item Our method achieves state-of-the-art performance among all camera-based SSC methods on the SemanticKITTI and OpenOccuapcny benchmarks.

\end{itemize}

\section{Related Work}

\subsection{3D Semantic Scene Completion}
Semantic Scene Completion (SSC), also known as semantic occupancy prediction, is a dense 3D perception task that jointly addresses semantic segmentation and scene completion~\cite{behley2019semantickitti,cai2021semantic}. 
Numerous previous works leverage LiDAR as the primary input to take advantage of the 3D geometrical information~\cite{zhang2018efficient,rist2021semantic,garbade2019two}.
Due to the cost-effectiveness and portability, camera-based 3D SSC is recently gaining increasing attention~\cite{behley2019semantickitti,silberman2012indoor,song2017semantic,straub2019replica,cai2021semantic,rist2021semantic,yan2021sparse,cheng2021s3cnet,wu2020scfusion,roldao2020lmscnet,li2020anisotropic,cao2022monoscene,li2023voxformer,huang2023tri}. MonoScene~\cite{cao2022monoscene} first proposes to infer geometry and semantics from a single RGB image with 2D-3D features projection. Inspired by this, a lot of the following works extend the domain of camera-based 3D scene perception~\cite{huang2023tri,zhang2023occformer,wei2023surroundocc,li2023stereoscene}. TPVFormer~\cite{huang2023tri} introduces a tri-perspective view to describe the fine-grained representation of a 3D scene. 
OccFormer~\cite{zhang2023occformer} introduces a dual-path transformer network to process dense 3D features for semantic occupancy prediction. 
SurroundOcc~\cite{wei2023surroundocc} proposes to estimate dense 3D occupancy with multi-view image inputs.
However, these methods attempt to describe the complicated 3D scene from single-timestep images, which is ineffective for such an inherently ill-posed problem due to incomplete visual cues.
In this paper, we advocate delving into reliable temporal content to dynamically aggregate semantic context and compensate for incomplete observations.\looseness=-1

\subsection{Temporal Information Modeling in 3D Visual Perception}
The utilization of temporal information is recently highlighted in temporal 3D object detection~\cite{zhang2019exploiting,zhang2019exploiting,wang2019unos,luo2020consistent,li2021enforcing,kopf2021robust,lin2022sparse4d,liu2022petr,liu2023petrv2} and video depth estimation~\cite{watson2021temporal,long2021multi,cai2023riav} to enhance prediction performance.
Temporal 3D object detection solutions focus on coarse-grained predictions at region-level~\cite{liu2023petrv2,lin2022sparse4d}, while video depth estimation methods~\cite{long2021multi,cai2023riav} aim to establish matching correspondence from sequential video frames.
Consequently, such strategies are insufficient for semantic scene completion, where fine-grained features are essential for dense semantic perception.
VoxFormer-T~\cite{li2023voxformer} builds the first temporal pipeline for camera-based SSC by simply stacking the features from different frames, 
while the temporal correspondence modeling for the dense perception task of SSC remains unexplored.
In this paper, we propose to explicitly model the temporal context correlation with pattern affinity to aggregate reliable temporal content and compensate for incomplete observations.

\begin{figure*}[!ht]
\vspace{-5pt}
\hsize=\textwidth %
\centering
\includegraphics[width=0.99\textwidth]{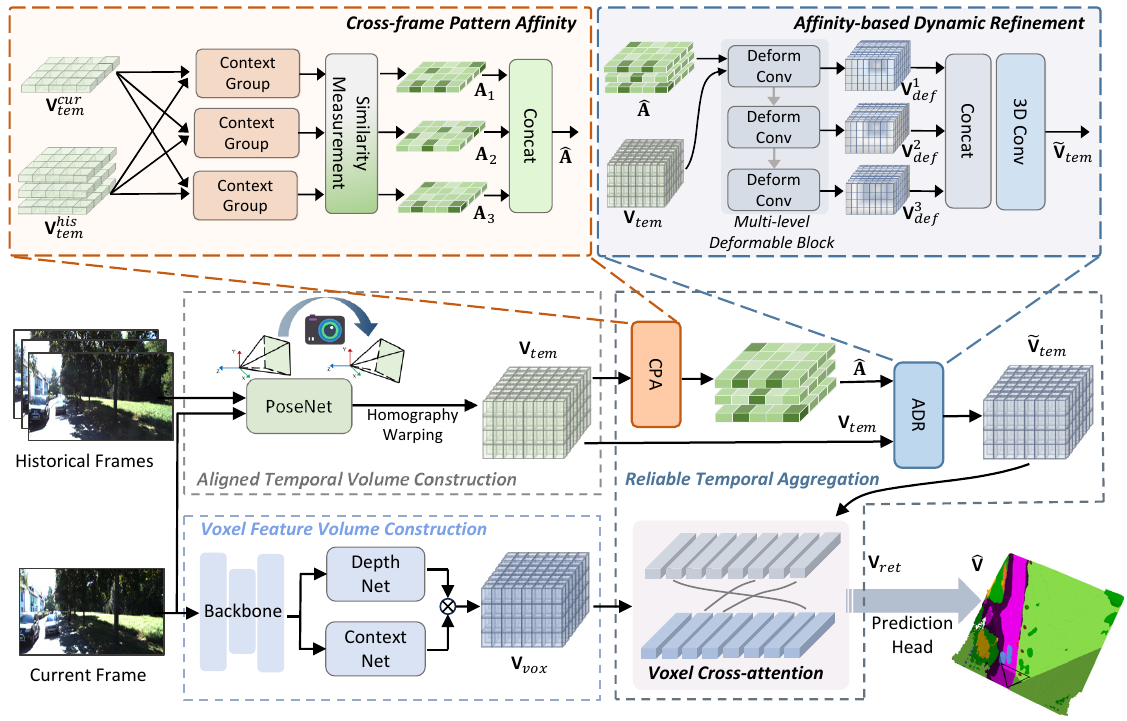}
\vspace{-12pt}
\caption{ {Overall framework of our proposed method}. 
Given temporal RGB images, the Aligned Temporal Volume is constructed with explicit epipolar homograph warping, while the Voxel Feature Volume is built by extending the LSS paradigm. 
Afterward, the Reliable Temporal Aggregation is introduced to dynamically aggregate reliable relevant temporal content for fine-grained semantic scene prediction. 
}
\label{figoverall}
 \vspace{-20pt}
\end{figure*}

\section{Methodology}

\subsection{Preliminary}

Given a set of input temporal RGB images $I_{{set}}^{rgb}=\{I^{rgb}_{t}, I^{rgb}_{t-1},\cdots \}$, our objective is to simultaneously estimate the semantic and geometric properties of the 3D scene.
Notably, our focus is exclusively on current and historical image frames, omitting consideration of future frames~\cite{li2023voxformer} to formulate a more practical scheme for real-world applications.
The scene is depicted as a voxel grid $\textbf{V}$ with dimensions $\mathbb{R}^{H\times W \times Z}$, where $H, W, Z$ denote the height, width and depth of the voxel grid, respectively. 
Each voxel within the grid is associated with a distinct semantic class represented by $C$, where $C$ takes values from the set $\{ c_0, c_1, \ldots, c_N \}$. The voxel can either correspond to empty space denoted as $c_0$ or to a particular semantic class from the set $\{c_1, c_2, \ldots, c_N\}$. Here, $N$ represents the total count of available semantic classes. 
Our objective is to leverage the proposed framework, denoted as $\Theta$, to learn a transformation:

\begin{equation}
\widehat{\textbf{V}} = \Theta ( I^{rgb}_{t}, I^{rgb}_{t-1},\cdots  ),
\end{equation}
where $\widehat{\textbf{V}}$ denotes the estimated 3D semantic voxel grid, aiming to approximate the ground truth voxel grid $\textbf{V}$.

\subsection{Overall Framework}
As depicted in Figure~\ref{figoverall}, the overall framework of our proposed method mainly consists of three components: Aligned Temporal Volume Construction in the upper branch, Voxel Feature Volume Construction in the lower branch, and Reliable Temporal Aggregation for fine-grained SSC prediction.

\noindent\textbf{Aligned Temporal Volume Construction.}
To construct the temporal feature volume $\textbf{V}_{tem}$, we feed the current and historical frames into a lightweight PoseNet~\cite{guizilini20203d,cai2023riav} to generate the temporal contextual volume $\textbf{V}_{tem}$ with homography warping. Different from computing matching costs in typical temporal depth estimation solutions~\cite{long2021multi,watson2021temporal,cai2023riav}, we advocate to maintain the context features with $\textbf{V}_{tem}$, further details are introduced in Section~\ref{sec_tca}.

\noindent\textbf{Voxel Feature Volume Construction.}
To construct the voxel feature volume $\textbf{V}_{vox}$, a UNet backbone based on pre-trained EfficientNetB7~\cite{tan2019efficientnet} is firstly employed to generate features with a spatial dimension of $\mathbb{R}^{H/4 \times W/4}$. Next, we extend the LSS~\cite{philion2020lift,li2023bevdepth} paradigm following recent studies~\cite{zhang2023occformer,li2023stereoscene} to build the voxel feature volume $\textbf{V}_{vox}$ from the outer product of the contextual information and the depth distribution.
To model the depth distribution, off-the-shelf monocular~\cite{bhat2021adabins} or stereo~\cite{cheng2020hierarchical} depth estimation networks are utilized with depth hypothesis planes of 192. By default, we employ the stereo depth estimation to form a stereo-based pipeline of \emph{HTCL-S}.
Moreover, we construct another monocular-based pipeline of \emph{HTCL-M} with the monocular depth estimation, broadening the applicability to scenarios without stereo inputs.

\noindent\textbf{Reliable Temporal Aggregation.}
We leverage the temporal volume $\textbf{V}_{tem}$ to form the cross-frame affinity $\hat{\textbf{A}}$, quantifying the contextual correspondence between current and historical features. Subsequently, we employ the cross-frame affinity to reassemble the temporal content and dynamically refine the sampling locations, yielding the reliable temporal volume $\widetilde{\textbf{V}}_{tem}$.
Further details on the Cross-frame Pattern Affinity (CPA) are introduced in Section~\ref{sec_mam}, while the Affinity-based Dynamic Refinement (ADR) is presented in Section~\ref{sec_dap}.

A Weighted Voxel Attention (WVA) is employed to aggregate reliable temporal content. 
Given $\textbf{V}_{vox}$ and $\widetilde{\textbf{V}}_{tem}$ as inputs, the query $Q$ is generated from $\textbf{V}_{vox}$, the key $K$ and the value $V$ are generated from $\widetilde{\textbf{V}}_{tem}$. 
During the early training phase, unregulated temporal information could impair the learning of the voxel feature volume.
To mitigate this, a flexible learning mechanism involving weighted voxel cross-attention is leveraged in the aggregation process:

\begin{equation} \label{eqca}
\begin{split}
\textbf{V}_{ret} &= \alpha \cdot \texttt{CrossAtt}(Q,K,V ) + \textbf{V}_{vox},
\end{split}
\end{equation}
where the learnable coefficient $\alpha$ is initialized with $0$ and gradually increases during training. $\textbf{V}_{ret}$ represents the aggregated volume, which is fed into an SSC head with upsampling and a softmax layer for SSC prediction $\widehat{\textbf{V}}$ following~\cite{cao2022monoscene,zhang2023occformer}.\looseness=-1

\subsection{Temporal Content Alignment}\label{sec_tca}
Given the fine-grained nature of the Semantic Scene Completion (SSC) task, constructing a dense temporal-aligned feature representation is crucial for accurate and robust perception.
Rather than simply stacking the input images from different viewpoints~\cite{li2023voxformer}, we propose to first align the temporal invariant content with explicit homography transformation.

As shown in Figure~\ref{figoverall}, the current and historical frames are first fed into a lightweight PoseNet following video depth estimation~\cite{guizilini20203d,watson2021temporal} to generate the relative camera pose for photometric reprojection.
Next, we leverage the current and historical frames to generate the current feature map $F_{t}$ and historical feature maps $\{ F_{t-1}, \cdots, F_{t-n} \}$. Following~\cite{cai2023riav,watson2021temporal}, we construct the warped historical features through homography warping with the relative camera pose and alternative depth hypothesis planes, which is formed as:

\begin{equation}
\mathrm{Warp}(\mathbf{p}) =\mathbf{K}_i \cdot\left(\mathbf{R}_{0, i} \cdot\left(\mathbf{K}_0{ }^{-1} \cdot \mathbf{p} \cdot d_j\right)+\mathbf{t}_{0, i}\right),
\end{equation}
where $\left\{\mathbf{K}_i\right\}_{i=0}^{N-1}$  and $\left\{\left[\mathbf{R}_{0, i} \mid \mathbf{t}_{0, i}\right]\right\}_{i=1}^{N-1}$ denote camera intrinsic parameters and extrinsic parameters, respectively.
$d_j$ denotes $j^{th}$ hypothesized depth of pixel $\mathbf{p}$ in $F_t$. 
Following that, we build a historical feature volume $\textbf{V}_{tem}^{his}$ by aggregating all warped historical features in the canonical space.
The historical feature volume contains geometric compatibility with different depth values between the current and historical frames.
Next, 
we lift $F_t$ along the depth dimension as~\cite{watson2021temporal,newcombe2011dtam} to generate the current feature volume $\textbf{V}_{tem}^{cur}$.
We concatenate $\textbf{V}_{tem}^{his}$ and $\textbf{V}_{tem}^{cur}$ following~\cite{watson2021temporal} to construct the temporal feature volume $\textbf{V}_{tem}$:
\begin{equation}
\begin{aligned}
\textbf{V}_{tem} &= \texttt{Concat} \left \{  (\textbf{V}_{tem}^{cur}, \textbf{V}_{tem}^{his}), \mathrm{dim}=\mathbb{C} \right \} \\
&=  \texttt{Concat} \left \{  \texttt{Lift}(F_t), \texttt{Warp}(F_{t-1}, \cdots, F_{t-n} ) \right \}.   \\
\end{aligned}
\end{equation}

The temporal volume $\textbf{V}_{tem}$ benefits the semantic scene modeling by explicitly aligning the contextual features across different timesteps. In the following section, we elaborate on fully exploiting reliable information according to contextual correspondence with $\textbf{V}_{tem}$. 

\noindent\textbf{Why feature volume instead of cost volume?}
The key distinction arises from the nature of camera-based Semantic Scene Completion, which is fundamentally not a matching task but rather a dense perception and reconstruction problem.
Consequently, instead of directly computing matching costs within the temporal feature volume, our approach prioritizes the maintenance of fine-grained feature context. 
{Moreover, to quantify the relevance of regional patterns within the temporal information, we construct auxiliary pattern affinity between the current and historical features.}

\subsection{Cross-frame Pattern Affinity Measurement}\label{sec_mam}

Although the temporal volume is explicitly aligned, it mixes redundant context from different frames, making it insufficient to directly model the scene representations corresponding to the current frame.
Therefore, we propose to construct Cross-frame Pattern Affinity (CPA) to measure the regional contextual correspondence between the historical feature volume $\textbf{V}_{his}$ and current feature volume $\textbf{V}_{cur}$.

\noindent\textbf{Similarity Measurement.}
As a classic similarity metric, cosine similarity is commonly used in semantic analysis~\cite{evangelopoulos2013latent,ramachandran2011automated,evangelopoulos2012latent} and information retrieval~\cite{rahutomo2012semantic,korenius2007principal} for correlation measurement. 
Given two vectors of $\alpha$ and $\beta$, the cosine similarity is calculated as:
\begin{equation}
\label{cos}
\texttt{sim}(\alpha, \beta)=\texttt{cos}(\vec{\alpha}, \vec{\beta})=\frac{\vec{\alpha} \cdot  \vec{\beta}}{||\vec{\alpha}|| \ast ||\vec{\beta}|| } . 
\end{equation}

However, the original cosine similarity may yield high similarity scores with two dissimilar vectors~\cite{sarwar2001item}.
This limitation is acknowledged and rectified through the scale-aware isolation~\cite{anastasiu2014l2ap}, which takes into consideration different pattern scales.
Nevertheless, these solutions tend to emphasize vector orientations and encounter challenges when assessing similarity within dense distributions. 
To overcome these drawbacks, ensemble learning techniques, as discussed in~\cite{xia2015learning}, 
leverage a diverse set of independent learners to address the aforementioned undesirable properties, thereby enhancing the effectiveness of dense similarity measurements.

Given these concerns, we identify the criteria of an optimal similarity measurement strategy for fine-grained representations in SSC: \emph{incorporation of diverse independent learning} and \emph{scale-aware isolation}.  
In pursuit of this objective, we advocate employing the scale-aware isolated cosine similarity and taking multi-group context as inputs for affinity computation with dense distributions. 
Our strategy is implemented through two key steps:
\begin{itemize}

    \item Incorporate different pattern scales from multi-group context to enable diverse independent similarity learning for fine-grained representations in SSC.
 
    \item Generate cosine similarities with scale-aware isolation and aggregate them for reliable pattern affinity measurement.

\end{itemize}

\noindent\textbf{Multi-group Context Generation.}
To facilitate diverse independent similarity learning, 3D atrous convolutions with different dilation rates are employed to construct multi-group contextual features.
Specifically, the historical feature volume $\textbf{V}_{tem}^{his}$ is processed by a set of atrous convolutions to generate historical multi-group context $\textbf{H}_i$ ($i \in ({1,2,3}$)):

\begin{equation}
\begin{aligned}
\textbf{H}_i =  \texttt{GN} \left ( \delta \left (  \texttt{Atrous}_i (\textbf{V}_{tem}^{his}) \right ) \right ) ,
\end{aligned}
\end{equation}
where $ \texttt{GN}$ and $\delta$ denote group normalization and GELU activation, respectively. 
The atrous convolutions are employed in parallel with dilation rates of 1, 2, and 4, respectively.
Note that the current multi-group context $\textbf{C}_i$ is generated in a symmetrical manner from the current feature volume $\textbf{V}_{tem}^{cur}$:
\begin{equation}
\begin{aligned}
\textbf{C}_i =  \texttt{GN} \left ( \delta \left (  \texttt{Atrous}_i (\textbf{V}_{tem}^{cur}) \right ) \right ) . 
\end{aligned}
\end{equation}

\begin{figure}[t]
\vspace{-5pt}
\centering
\includegraphics[width=8.5cm]{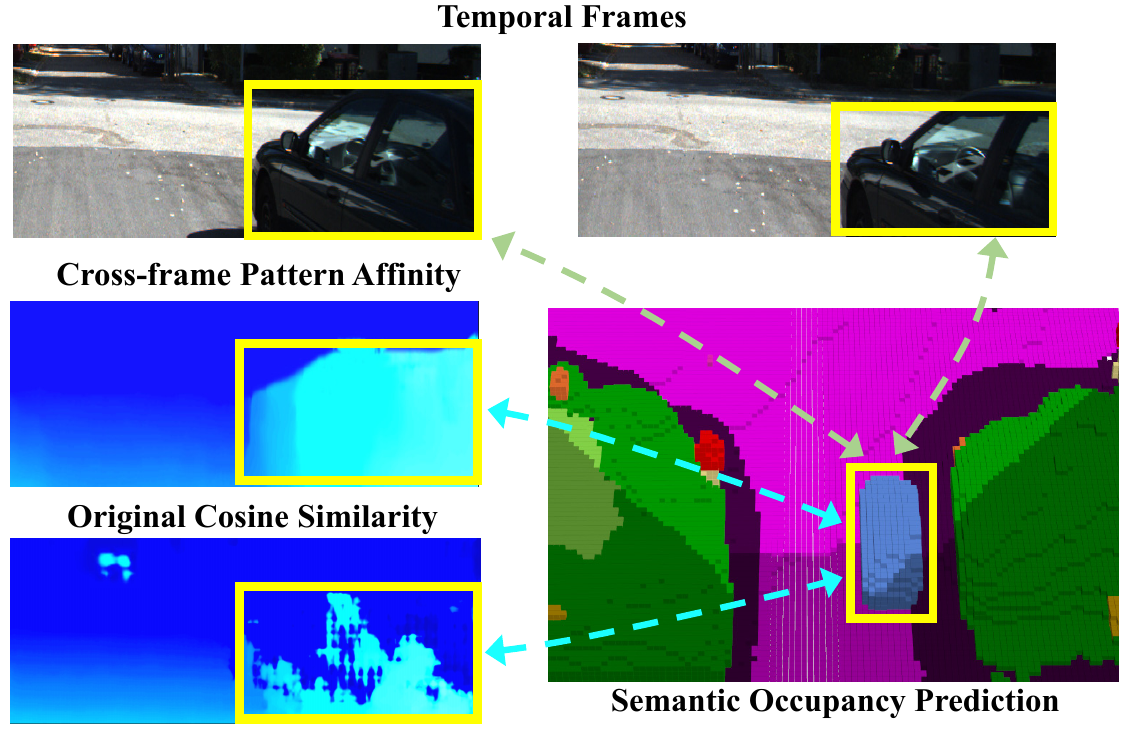}  
\vspace{-5pt}
\caption{Visualization of the heat maps from our proposed Cross-frame Pattern Affinity (CPA) and the original cosine similarity. 
} 
\label{fig_aff}
\vspace{-15pt}
\end{figure} 

\noindent\textbf{Measure Pattern Affinity for Dense SSC.} 
We upgrade Equation~\ref{cos} with two primary modifications to formulate the pattern affinity measurement for fine-grained contextual correspondence modeling in SSC. Firstly, we consider different pattern scales of the multi-group context and compute the pattern affinity $\textbf{A}_{i}$ with each scale $i$. These independent group-scale affinity matrices are further aggregated along the channel dimension.
Secondly, we subtract the respective averages during the affinity computation within each group scale to achieve scale-aware isolation. The formulas are represented as:

\begin{align}
\textbf{A}_{i}  &=  \texttt{sim} ( \textbf{C}_{i}, \textbf{H}_{i}) \\ \notag
 &=  \frac{ \sum_{j=0}^{C} ( \textbf{C}_{i}^{j} - \overline{ \textbf{C}}_{i} )(\textbf{H}^{j}_{i} - \overline{ \textbf{H}}_{i})}{ \sqrt{ {\sum_{j=0}^{C} ( \textbf{C}^{j}_{i} - \overline{ \textbf{C}}_{i} )}^{2} } \sqrt{ {\sum_{j=0}^{C} ( \textbf{H}^{j}_{i} - \overline{ \textbf{H}}_{i} )}^{2} } } , \\ 
 \hat{\textbf{A}} &=  \texttt{Concat}  \left \{ (\textbf{A}_1,\textbf{A}_2,\textbf{A}_3), \mathrm{dim}=\mathbb{C}  \right \} , 
\end{align}
where the affinity matrices $\textbf{A}_{i}$ of different group scales are concatenated along the channel dimension to get the cross-frame pattern affinity $\hat{\textbf{A}}$. 
The input context matrices $\textbf{C}_{i}$ and $\textbf{H}_{i}$ are taken as high-dimension vectors with different group scales. 
$\overline{ \textbf{C}}_{i}$ and $\overline{ \textbf{H}}_{i}$ represent averaged context matrices of each group scale. 
As illustrated in Figure~\ref{fig_aff}, 
the affinity map from Cross-frame Pattern Affinity (CPA) effectively signifies contextual correspondence within the temporal content.

\subsection{Affinity-based Dynamic Refinement}\label{sec_dap}

Given our objective of completing and comprehending the 3D scene corresponding to the current frame, it is essential to assign greater weights to the most relevant locations.
Simultaneously, investigating their neighboring relevant context is also critical to compensate for incomplete observations.

To this end, we propose to 
adaptively refine the feature sampling locations based on the obtained identified high-affinity locations and their neighboring relevant regions.
We implement the above ideas with 3D deformable convolutions~\cite{dai2017deformable,wang2021patchmatchnet}.
Specifically, we accomplish the dynamic refinement by introducing the affinity-based correspondence weights and deformable position offsets.
In the context of a sampling grid window $K_w$, the formula is expressed as:

\begin{equation} \label{eqca2}
\begin{split}
\textbf{V}_{def} = \sum_{k=1}^{K_w} w_k \cdot   \textbf{V}_{tem}(\textbf{p}+\textbf{p}_k+\Delta \textbf{p}_k) \cdot a_{k},
\end{split}
\end{equation}
where $K_w$ represents the number of points in the sampling process. $\Delta \textbf{p}_k$ denotes the additional offset in the sampling grid. $w_k$ denotes the spatial feature weight and $a_{k}$ represents the affinity weight from the cross-frame pattern affinity $\hat{\textbf{A}}$.

To further reason about dynamic modeling through hierarchical context, we optimize the refinement process by considering contextual information from different feature levels. As illustrated in Figure~\ref{figoverall}, we construct a multi-level deformable block with three cascade 3D deformable convolutions. The output features are aggregated to generate the reliable temporal volume $\widetilde{\textbf{V}}_{tem}$:

\begin{equation} \label{eqca3}
\begin{split}
\widetilde{\textbf{V}}_{tem} &= \textbf{W}  \left ( \texttt{Concat} \left \{ ( \textbf{V}_{def}^{1},\textbf{V}_{def}^{2},\textbf{V}_{def}^{3}  ) , \mathrm{dim}=\mathbb{C}  \right \}  \right ) .
\end{split}
\end{equation}
where the multi-level deformable temporal volumes $\textbf{V}_{def}^{i} \left ( i \in \left ( 1,2,3\right ) \right )$ are concatenated along the channel dimension and processed with a 3D convolution layer $\textbf{W}$ for dimension reduction.

\section{Experiment}

\subsection{Datasets and Metrics}
\noindent\textbf{SemanticKITTI.}
The SemanticKITTI~\cite{behley2019semantickitti} dataset comprises 22 outdoor scenes with LiDAR scans and stereo images. 
The ground truth is voxelized as 256$\times$256$\times$32 grids. Each voxel grid has a size of ($0.2m$, $0.2m$, $0.2m$) and is annotated with 21 semantic classes (19 semantics, 1 free and 1 unknown). 
Following~\cite{cao2022monoscene,li2023voxformer}, we divide the 22 outdoor scenes into 10 training scenes, 1 validation scene, and 1 test scene.
Our proposed HTCL is evaluated on the SemanticKITTI with both temporal stereo images (HTCL-S) and monocular images (HTCL-M).

\noindent\textbf{OpenOccupancy.}
The OpenOccupancy~\cite{wang2023openoccupancy} dataset extends the nuScene~\cite{caesar2020nuscenes} dataset by providing dense semantic occupancy annotations. The dataset holds 850 scenes of 34K keyframes with 360-degree LiDAR scans. We split the whole dataset into 28130 training frames and 6019 validation frames following~\cite{wang2023openoccupancy}. Each frame holds 400K occupied voxels with 17 semantic labels.
Note that we exclusively apply monocular-based HTCL-M on the OpenOccupancy dataset due to the unavailability of stereo images.

\noindent \textbf{Evaluation Metrics.}
Following the previous works~\cite{cao2022monoscene,li2023voxformer}, we utilize the mean Intersection over Union \textbf{(mIoU)} as our primary metric to assess the performance of the semantic scene completion (SSC) task. Additionally, we report the Intersection over Union \textbf{(IoU)} to evaluate the performance of the class-agnostic scene completion (SC) task.

\subsection{Experimental Setup}\label{version_fast}
We follow the common practice~\cite{cao2022monoscene,zhang2023occformer,li2023voxformer} to initialize the encoder part of our UNet with the pretrained weight of EfficientNetB7~\cite{tan2019efficientnet}.
By default, our model takes the current frame and the previous 3 image frames as inputs.
We implemented our model on PyTorch with a batch size of 4. The model is trained 24 epochs with the AdamW optimizer~\cite{loshchilov2017decoupled}. The learning rate is set to $1\times10^{-4}$ with a weight decay of 0.01.

\begin{table*}[!ht]
\vspace{-5pt}
\begin{center}
\scriptsize
\caption{\textbf{Quantitative results} on the SemanticKITTI test set with the state-of-the-art SSC methods. 
The {``S-T'', ``S''} and {``M''} denote temporal stereo images, single-frame stereo images, and single-frame monocular images, respectively.  
}
\vspace{-10pt}
\renewcommand\tabcolsep{1.1pt}
\resizebox{0.99\textwidth}{!}{%
\begin{tabular}{l|c|c|c|c|c|c|c}
\toprule
{Methods}   & HTCL-S (ours)    & VoxFormer-T  & VoxFormer-S  & OccFormer &SurroundOcc & TPVFormer  & MonoScene  \\ \midrule

{Input}     & {S-T}  &{S-T} &{S} &{M} &{M} &{M} & {M}  \\ \midrule
\textbf{IoU }  & \textbf{44.23}   & {43.21} &42.95 & 34.53  & 34.72 &34.25 &34.16      \\ \midrule
\textbf{mIoU }      & \textbf{17.09}  & {13.41}  &12.20 &12.32 & 11.86 &11.26   &11.08      \\ \midrule

\crule[carcolor]{0.13cm}{0.13cm} {car}  & \textbf{27.30}  & {21.70} & 20.80 & 21.60 &20.60 & 19.20   &  18.80  \\

\crule[bicyclecolor]{0.13cm}{0.13cm} {bicycle}  & {1.80} &\textbf{1.90} & 1.00& 1.50 &1.60 & 1.00 & {0.50} \\

\crule[motorcyclecolor]{0.13cm}{0.13cm} {motorcycle}    & \textbf{2.20} &{1.60} &0.70 & {1.70} &1.20 & 0.50  &  {0.70}   \\

\crule[truckcolor]{0.13cm}{0.13cm} {truck}   & \textbf{5.70} & {3.60} &3.50 & {1.20} &1.40  & {3.70}  &  {3.30}   \\

\crule[othervehiclecolor]{0.13cm}{0.13cm} {other-veh.} & \textbf{5.40}  &4.10 & 3.70 & 3.20 &{4.40} & 2.30  &{4.40}   \\

\crule[personcolor]{0.13cm}{0.13cm} {person}  & {1.10} &{1.60} &1.40& \textbf{2.20} &1.40 &{1.10}  & {1.00} \\

\crule[bicyclistcolor]{0.13cm}{0.13cm} {bicyclist}  & \textbf{3.10}  & 1.10 & {2.60} &1.10 &2.00  & {2.40} & {1.40}   \\

\crule[motorcyclistcolor]{0.13cm}{0.13cm} {motorcyclist} & \textbf{0.90}  & 0.00 &0.20 & 0.20 &0.10 & 0.30   &  {0.40}  \\

\crule[roadcolor]{0.13cm}{0.13cm} {road}  & \textbf{64.40}   &54.10 &53.90 & 55.90 &{56.90} &{55.10}    &{54.70}  \\

\crule[parkingcolor]{0.13cm}{0.13cm} {parking}  & \textbf{33.80} & 25.10& 21.10 &{31.50} &30.20 & {27.40}   & {24.80}  \\

\crule[sidewalkcolor]{0.13cm}{0.13cm} {sidewalk}    & \textbf{34.80}  &26.90 &25.30 & {30.30} &28.30 & {27.20} & {27.10}   \\

\crule[othergroundcolor]{0.13cm}{0.13cm} {other.grd} & \textbf{12.40}  & {7.30} &5.60 &6.50 &6.80 & 6.50 & {5.70}    \\

\crule[buildingcolor]{0.13cm}{0.13cm} {building}    & \textbf{25.90}  &{23.50} & 19.80 &15.70 &15.20 & 14.80 & {14.40}  \\

\crule[fencecolor]{0.13cm}{0.13cm} {fence}  & \textbf{21.10}  &{13.10} & 11.10 & 11.90 & 11.30 & {11.00} &  {11.10} \\

\crule[vegetationcolor]{0.13cm}{0.13cm} {vegetation}   & \textbf{25.30} & {24.40} & 22.40 & 16.80 &14.90 & 13.90 & {14.90}  \\

\crule[trunkcolor]{0.13cm}{0.13cm} {trunk}  & \textbf{10.80} & {8.10} & 7.50 & 3.90 & 3.40 & 2.60   &2.40\\

\crule[terraincolor]{0.13cm}{0.13cm} {terrain}   & \textbf{31.20} & {24.20} &21.30 & 21.30 &19.30  &20.40 & {19.50}   \\

\crule[polecolor]{0.13cm}{0.13cm} {pole}  & \textbf{9.00}   & {6.60} & 5.10 & 3.80 &3.90 & 2.90 &3.30   \\

\crule[trafficsigncolor]{0.13cm}{0.13cm} {traf.sign}   & \textbf{8.30}  &{5.70} & 4.90 & 3.70 &2.40 & 1.50 &2.10  \\  \bottomrule

\end{tabular}
}
\label{tabq1}        
\vspace{-10pt}
\end{center}
\end{table*}

\begin{table*}[!ht]
\vspace{-0pt}
\begin{center}
\scriptsize
\caption{\textbf{Quantitative results} on the OpenOccupancy validation set with the state-of-the-art SSC methods. The ``L'', ``M'', ``M-D'' and ``M-T'' denote LiDAR inputs, monocular images, monocular images with depth maps and temporal monocular images, respectively. The LiDAR points are projected and densified to generate the depth maps.}
\vspace{-0pt}
\renewcommand\tabcolsep{2.6pt}
\resizebox{0.96\textwidth}{!}{%
\begin{tabular}{l|c|c|c|c|c|c|c}
\toprule
{Methods}   & HTCL-M (ours) & JS3C-Net & LMSCNet  & 3DSketch & AICNet & TPVFormer  & MonoScene  \\ \midrule

{Input}     & {M-T}  &{L} &{L} &{M-D} &{M-D} &{M} & {M}  \\ \midrule
\textbf{IoU}  & 21.4  &\textbf{30.2} & 27.3& 25.6& 23.8 & 15.3 & 18.4  \\ \midrule
\textbf{mIoU} &\textbf{14.1} & 12.5  & 11.5& 10.7 & 10.6 & 7.8 & 6.9 \\ \midrule

\rotatebox{0}{\textcolor{barrier1}{$\blacksquare$} {barrier}}  & \textbf{14.8} & 14.2 & 12.4 & 12.0& 11.5 & 9.3 & 7.1  \\

\rotatebox{0}{\textcolor{bicycle1}{$\blacksquare$} {bicycle}} &\textbf{10.2} & 3.4& 4.2 &5.1 & 4.0 & 4.1 & 3.9\\

\rotatebox{0}{\textcolor{bus1}{$\blacksquare$} {bus}}  &\textbf{14.8} & 13.6 & 12.8 &10.7& 11.8&11.3 &9.3\\

\rotatebox{0}{\textcolor{car1}{$\blacksquare$} {car}}  &\textbf{18.9} & 12.0& 12.1 &12.4 & 12.3 &10.1 &7.2\\

\rotatebox{0}{\textcolor{const. veh.1}{$\blacksquare$} {const. veh.}}  &\textbf{7.6} & 7.2& 6.2 &6.5 &5.1 &5.2 &5.6\\

\rotatebox{0}{\textcolor{motorcycle1}{$\blacksquare$} {motorcycle}} &\textbf{11.3}  &4.3 &4.7 &4.0& 3.8 & 4.3 &3.0\\

\rotatebox{0}{\textcolor{pedestrian1}{$\blacksquare$} {pedestrian}} &\textbf{12.3}& 7.3& 6.2 &5.0& 6.2 &5.9  &5.9\\

\rotatebox{0}{\textcolor{traffic cone1}{$\blacksquare$} {traffic cone}} &\textbf{9.6}& 6.8& 6.3 &6.3& 6.0 & 5.3 & 4.4\\

\rotatebox{0}{\textcolor{trailer1}{$\blacksquare$} {trailer}} & 5.5 &\textbf{9.2}& 8.8&8.0 & 8.2 &  6.8 & 4.9\\

\rotatebox{0}{\textcolor{truck1}{$\blacksquare$} {truck}} &\textbf{13.5}& 9.1& 7.2&7.2 &7.5 & 6.5 & 4.2\\

\rotatebox{0}{\textcolor{drive. suf.1}{$\blacksquare$} {drive. suf.}} &\textbf{32.5} & 27.9& 24.2&21.8 &24.1 & 13.6 & 14.9 \\

\rotatebox{0}{\textcolor{other flat1}{$\blacksquare$} {other flat}} &\textbf{21.7} & 15.3& 12.3 &14.8 & 13.0 & 9.0 & 6.3 \\

\rotatebox{0}{\textcolor{sidewalk1}{$\blacksquare$} {sidewalk}} &\textbf{20.7} & 14.9& 16.6 &13.0 & 12.8& 8.3 & 7.9 \\

\rotatebox{0}{\textcolor{terrain1}{$\blacksquare$} {terrain}} &\textbf{17.7} & 16.2& 14.1 &11.8 & 11.5& 8.0 & 7.4 \\

\rotatebox{0}{\textcolor{manmade1}{$\blacksquare$} {manmade}} & 5.8 & \textbf{14.0}& 13.9 & 12.0 & 11.6& 9.2 & 10.0 \\

\rotatebox{0}{\textcolor{vegetation1}{$\blacksquare$} {vegetation}}  & 8.5 & \textbf{24.9}& 22.2 & 21.2 & 20.2& 8.2 & 7.6\\
 \bottomrule
\end{tabular}
}
\vspace{-0pt}
\label{tab_open}
\end{center}
\end{table*}

\subsection{Performance.}

\textbf{Quantitative Comparison.}
As reported in Table~\ref{tabq1}, we compare our HTCL with recent public methods on the SemanticKITTI dataset, including VoxFormer~\cite{li2023voxformer}, OccFormer~\cite{zhang2023occformer}, SurroundOcc~\cite{wei2023surroundocc}, TPVFormer~\cite{huang2023tri} and MonoScene~\cite{cao2022monoscene}.
VoxFomer-T is a temporal baseline with current and historical 4 images as inputs.
We can observe that our proposed method outperforms all the other methods significantly.
Compared to VoxFomer-T, our method achieves a remarkable relative improvement 
in mIoU even with fewer historical inputs (3~vs.~4).
We also report quantitative results on the OpenOccupancy validation set in Table~\ref{tab_open}.
To provide depth maps for AICNet~\cite{li2020anisotropic} and 3DSketch~\cite{3d-sketch}, LiDAR points are projected and densified following OpenOccupancy~\cite{wang2023openoccupancy}.
Despite LiDAR’s inherent advantage in terms of IoU due to more accurate 3D geometric measurement, our HTCL surpasses all the other methods (including LiDAR-based LMSCNet~\cite{roldao2020lmscnet} and JS3C-Net~\cite{yan2021sparse}) in terms of mIoU, which demonstrates the effectiveness of our method for semantic scene completion.

 \begin{figure*}[t]
   \vspace{-5pt}
	\begin{center}
		\includegraphics[width=12.0
  cm]{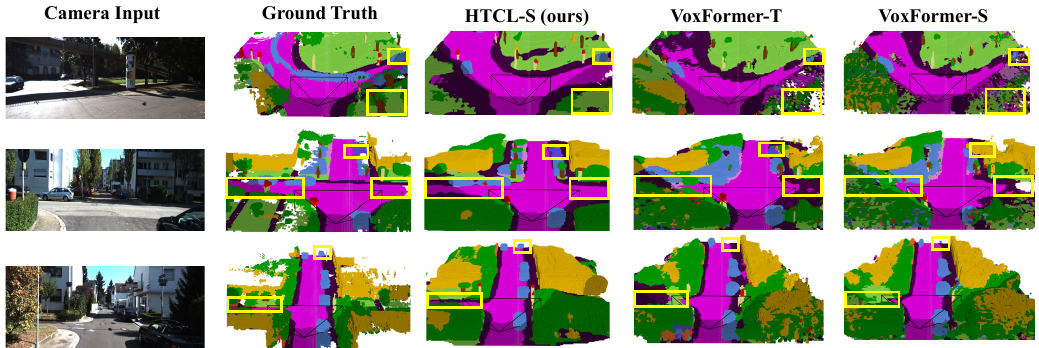}   
        \vspace{0pt}
		\begin{tabular}{cccccc}
			\multicolumn{6}{c}{
				\scriptsize
				\textcolor{bicycle}{$\blacksquare$}bicycle~
				\textcolor{car}{$\blacksquare$}car~
				\textcolor{motorcycle}{$\blacksquare$}motorcycle~
				\textcolor{truck}{$\blacksquare$}truck~
				\textcolor{other-vehicle}{$\blacksquare$}other-veh.~
				\textcolor{person}{$\blacksquare$}person~
				\textcolor{bicyclist}{$\blacksquare$}bicyclist~
				\textcolor{motorcyclist}{$\blacksquare$}motorcyclist~
				\textcolor{road}{$\blacksquare$}road~
				}
    \\
			\multicolumn{6}{c}{
				\scriptsize
                    \textcolor{parking}{$\blacksquare$}parking \textcolor{sidewalk}{$\blacksquare$}sidewalk \textcolor{other-ground}{$\blacksquare$}other.grd \textcolor{building}{$\blacksquare$}building \textcolor{fence} {$\blacksquare$}fence \textcolor{vegetation}{$\blacksquare$}vegetation \textcolor{trunk}{$\blacksquare$}trunk \textcolor{terrain}{$\blacksquare$}terrain \textcolor{pole}{$\blacksquare$}pole \textcolor{traffic-sign}{$\blacksquare$}traf.sign	
			}
		\end{tabular}	
 	\end{center}
        \vspace{0pt}
	\caption{ \textbf{Qualitative results} on the SemanticKITTI validation set. Our proposed HTCL captures more complete and accurate scenery layouts compared with VoxFormer. Meanwhile, HTCL hallucinates more proper scenery beyond the camera field of view.} 
	\label{fig_q}
 \vspace{-0pt}
\end{figure*}

\begin{figure}[t]
   \vspace{-0pt}
	\begin{center}
		\includegraphics[width=9
  cm]{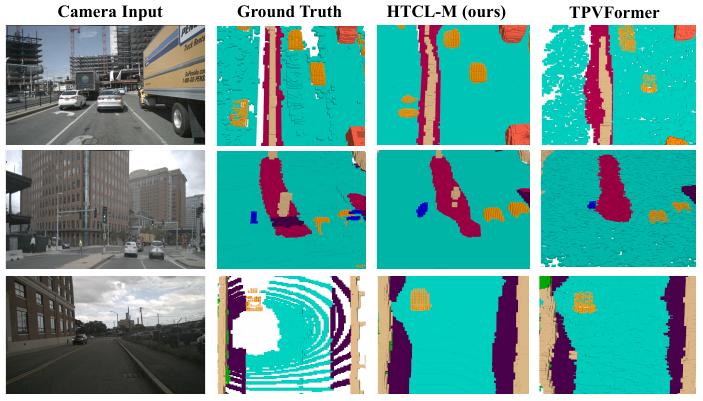}   
        \vspace{0pt}
		\begin{tabular}{cccccccc}
			\multicolumn{8}{c}{
				\scriptsize
				\textcolor{bicycle1}{$\blacksquare$}bicycle \textcolor{bus1}{$\blacksquare$}bus \textcolor{car1}{$\blacksquare$}car \textcolor{const. veh.1}{$\blacksquare$}const.veh. \textcolor{motorcycle1}{$\blacksquare$}motorcycle \textcolor{pedestrian1}{$\blacksquare$}pedestrian \textcolor{truck1}{$\blacksquare$}truck \textcolor{sidewalk1}{$\blacksquare$}sidewalk~
        
				}
    \\
			\multicolumn{7}{c}{
				\scriptsize

                \textcolor{traffic cone1}{$\blacksquare$}traffic cone \textcolor{trailer1}{$\blacksquare$}trailer \textcolor{drive. suf.1}{$\blacksquare$}drive.suf. \textcolor{other flat1}{$\blacksquare$}other flat \textcolor{terrain1}{$\blacksquare$}terrain \textcolor{manmade1}{$\blacksquare$}manmade \textcolor{vegetation1}{$\blacksquare$}vegetation~
			}
		\end{tabular}	
 	\end{center}
        \vspace{-17pt}
	\caption{ \textbf{Qualitative results} on the OpenOccupancy validation set. 
 Our proposed HTCL-M can generate more complete and comprehensive semantic scenes
 compared with the ground truth.
 } 
	\label{fig_open}
 \vspace{-20pt}
\end{figure}

\noindent \textbf{Qualitative Comparison.} 
Figure~\ref{fig_q} provides a qualitative comparison between our proposed method and VoxFormer on the SemanticKITTI validation set. 
We can observe that the real-world scenes are complex and the annotated ground truth is relatively sparse, which poses challenges for completely reconstructing the semantic scenes from limited visual cues.
Compared with VoxFormer, our method captures a more complete and accurate scenery layout (e.g., crossroads in the second and third rows).
Moreover, our method effectively hallucinates more proper scenery outside of the camera field of view (e.g., shadow areas in the first and second rows) and demonstrates significant superiority over moving objects (e.g., trucks in the second row). 
We also visualize the prediction results of our proposed method on the OpenOccupancy validation set as shown in Figure~\ref{fig_open}. Our proposed method
generates much denser and more realistic results compared with the ground truth.

\begin{table}[t]
\vspace{-0pt}
\begin{center}
\renewcommand\tabcolsep{18.0pt}
\scriptsize
\caption{ \textbf{Evaluation results} of temporal stereo variants.  
For MonoScene${}^{\ddagger}$, TPVFormer${}^{\ddagger}$ and OccFormer${}^{\ddagger}$, we employ stacked temporal stereo images as inputs following VoxFormer-T.}
\vspace{-10pt}
\begin{tabular}{l|ccc} 
\toprule
 \textbf{Methods}& \textbf{Input}   & \textbf{mIoU (\%)} $\uparrow$ & \textbf{Time (s)} $\downarrow$      \\  \midrule

MonoScene${}^{\ddagger}$ (2022) &  Stereo-T  & 12.96 & \textbf{0.281}  \\
TPVFormer${}^{\ddagger}$ (2023)   &  Stereo-T   &   13.21 & 0.324   \\
OccFormer${}^{\ddagger}$ (2023)  & Stereo-T  &  13.57 & 0.348   \\

VoxFormer-T (2023)  &  Stereo-T  & 13.35 & 0.307    \\ 
\rowcolor{gray!10} HTCL-M (ours) & Mono-T & 16.11  & 0.289  \\ 
\rowcolor{gray!10} HTCL-S (ours) & Stereo-T &\textbf{17.13} & {0.297}   \\ 
 \bottomrule
\end{tabular}
\vspace{-0pt}
\label{tabq3}
\end{center}
\end{table}

\noindent \textbf{Temporal Stereo Variants Evaluation.}
To ensure a fair and comprehensive comparison, we implement temporal stereo variants of the baselines as shown in Table~\ref{tabq3}. Following VoxFormer-T, we employ stacked temporal stereo images as inputs to get variants of MonoScene${}^{\ddagger}$, TPVFormer${}^{\ddagger}$ and OccFormer${}^{\ddagger}$. 
Note that VoxFormer-T originally employs 4 previous frames, while the other stereo variants employ 3 previous frames as ours.
As shown in the table, our method efficiently achieves superior performance with the same temporal inputs.

\begin{table*}[t]\centering
\vspace{-5pt}
\renewcommand\tabcolsep{3.0pt}
\scriptsize
\caption{{Ablation study for different architectural components} on the SemanticKITTI validation set. The full names of different components are in Sec~\ref{sec_ab}.}
\vspace{-12pt}
\resizebox{0.98\textwidth}{!}{
\begin{tabular}{cc|cc|cc|cc|cc} \toprule
\multicolumn{2}{c|}{\textbf{TCA}}                     & \multicolumn{2}{c|}{\textbf{CPA}}                     & \multicolumn{2}{c|}{\textbf{ADR}}                     & \multicolumn{2}{c|}{\textbf{WVA}} & 
\multirow{2}{*}{\textbf{IoU (\%)}} & \multirow{2}{*}{\textbf{mIoU (\%)}} \\
\begin{tabular}[c]{@{}c@{}}Feature\\Volume\end{tabular}
  & \begin{tabular}[c]{@{}c@{}}Cost\\Volume\end{tabular}   & \begin{tabular}[c]{@{}c@{}}Scale-aware \\Isolation\end{tabular}   & \begin{tabular}[c]{@{}c@{}}Multi-\\group\end{tabular}   & Affinity & Deformable  & Coefficient  &  CrossAtt &   \\ \midrule
 &\checkmark  &\checkmark &\checkmark &\checkmark & \checkmark &\checkmark & \checkmark &  44.01  &16.02  \\ \midrule

\checkmark &  & &\checkmark & \checkmark &\checkmark &\checkmark & \checkmark &43.07  &  15.18 \\
\checkmark &  & \checkmark& & \checkmark &\checkmark &\checkmark  & \checkmark & 43.15 &  15.26\\  \midrule

\checkmark &  &\checkmark &\checkmark &  & \checkmark&\checkmark & \checkmark & 42.79 & 14.65\\
\checkmark &  &\checkmark &\checkmark & \checkmark & &\checkmark  & \checkmark & 42.96 & 15.14 \\  \midrule
\checkmark &  &\checkmark &\checkmark &\checkmark  &\checkmark &   & \checkmark & 43.97 & 15.85 \\ 
\checkmark &  &\checkmark &\checkmark &\checkmark  &\checkmark & \checkmark  &  & 44.13 & 15.98 \\  \midrule
\rowcolor{gray!10}  \checkmark & &\checkmark &\checkmark &\checkmark & \checkmark &\checkmark & \checkmark & \textbf{45.51} & \textbf{17.13}  \\ \bottomrule
\end{tabular}
}
\vspace{-10pt}
\label{tab_ar}
\end{table*}

\subsection{Ablation Study}\label{sec_ab}
\begin{figure}[!ht]
 \vspace{-18pt}
\begin{minipage}[b]{0.5\linewidth}
\begin{center}
\captionsetup{type=table}
\caption{{Effect of using a different number of temporal frames.} These models are evaluated on the SemanticKITTI validation set.}
 \vspace{5pt}
\renewcommand\tabcolsep{3.0pt}
\renewcommand{\arraystretch}{1.5}
\scriptsize
\resizebox{1.0\textwidth}{!}{
\begin{tabular}{ccccc|cc}\toprule
\multicolumn{5}{c|}{\textbf{Temporal Inputs}} &\multirow{2}{*}{\textbf{mIoU (\%)} $\uparrow$} &\multirow{2}{*}{\textbf{Time (s)} $\downarrow$}   \\
 $I^{rgb}_{t-1}$ & $I^{rgb}_{t-2}$ & $I^{rgb}_{t-3}$ &$I^{rgb}_{t-4}$ &$I^{rgb}_{t-5}$     \\ \midrule
\checkmark & &  &  & & 15.08 & 0.268   \\
 \checkmark&\checkmark &  & & & 16.43 & 0.283 \\
\rowcolor{gray!10}  \checkmark  & \checkmark &\checkmark &&  &  17.13  & 0.297   \\
 \checkmark  & \checkmark &\checkmark &\checkmark&     & 17.31  & 0.311 \\
 \checkmark  & \checkmark &\checkmark &\checkmark &\checkmark  & 17.42 & 0.324 \\ \bottomrule
\end{tabular}
}
\vspace{0pt}
\label{tab_tem}
\end{center}
\end{minipage}
\hfill
\begin{minipage}[b]{0.45\linewidth}
\begin{center}
\includegraphics[width=1.0\linewidth]{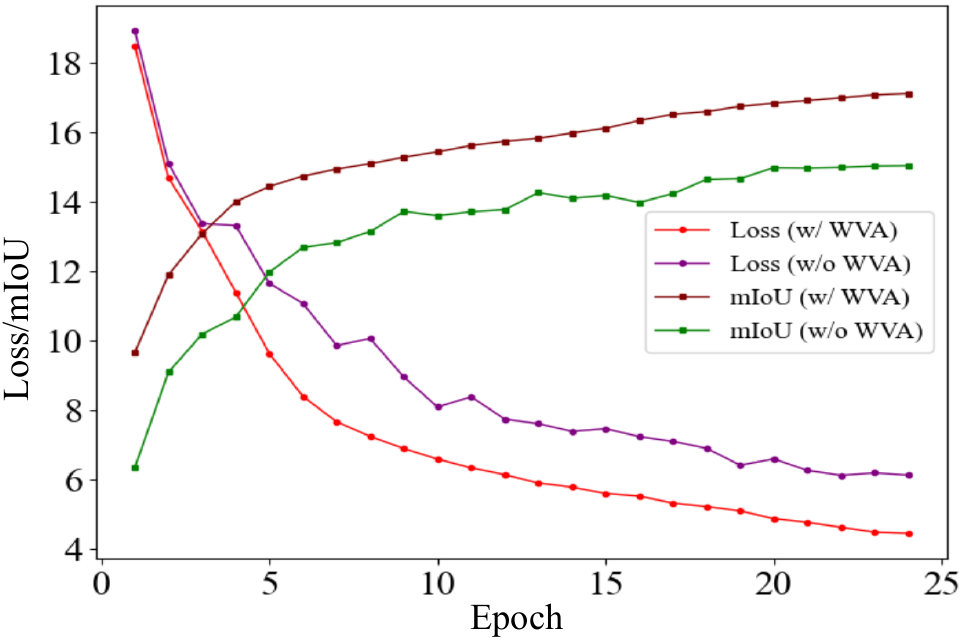}
 \vspace{-20pt}
\caption{{The convergence and performance curves} of WVA.}
\label{fig_wva}
\end{center}
\end{minipage}
 \vspace{-15pt}
\end{figure}

We conduct adequate ablations for our proposed method on the SemanticKITTI validation
set. Specifically, we analyze the impact of different architectural components in Table~\ref{tab_ar} and study the influence of temporal inputs in Table~\ref{tab_tem}.

\noindent \textbf{Temporal Content Alignment (TCA).}
The ablation study for Temporal Content Alignment (TCA) is reported 
in the second row of Table~\ref{tab_ar}. We can observe that replacing the cost volume with the feature volume yields obvious performance gains, improving the IoU and mIoU by 1.50 and 1.11, respectively. We attribute this enhancement to the fine-grained feature context preservation.

\noindent \textbf{Cross-frame Pattern Affinity (CPA).}
The ablation of Cross-frame Pattern Affinity (CPA) is detailed in the third row of Table~\ref{tab_ar}. As we can see, equipping the original cosine similarity with the scale-aware isolation and introducing the multi-group context generation lead to significant performance enhancement, improving the mIoU by 1.95 and 1.87, respectively. 

\noindent \textbf{Affinity-based Dynamic Refinement (ADR).}
The ablation study for the Affinity-based Dynamic Refinement (ADR) is conducted by removing the affinity weights and replacing the deformable convolutions with normal convolutions, as detailed in the fourth row of Table~\ref{tab_ar}. Leveraging the affinity information is effective in modeling contextual correspondence, as the procedure results in a noticeable performance gain of 2.72 IoU and 2.48 mIoU. Furthermore, dynamic refinement with the deformable convolutions offers efficient and flexible contextual modeling, improving IoU and mIoU by 2.55 and 1.99, respectively.

\noindent \textbf{Weighted Voxel Attention (WVA).}
The ablation study about Weighted Voxel Attention (WVA) is shown in Figure~\ref{fig_wva} and the fifth row of Table~\ref{tab_ar}. We remove the learnable coefficient and replace the voxel cross-attention with naive concatenation for comparison. 
It is evident in Figure~\ref{fig_wva} that the adoption of our proposed strategy leads to a more rapid and stable convergence of the entire model. Additionally, Table~\ref{tab_ar} shows noteworthy improvements through the incorporation of the learnable coefficient and the voxel cross-attention, enhancing the mIoU by 1.28 and 1.15, respectively.

\noindent \textbf{Temporal Inputs.}
We report the semantic scene completion performance and running time with temporal inputs of different frame numbers, which are detailed in Table~\ref{tab_tem}. 
As we can observe, the effectiveness gain of more than 3 previous frames is relatively slight with more running time, thus we adopt 3 frames as the default setting to balance between efficiency and effectiveness.

\vspace{-0pt}
\section{Conclusion}
\vspace{-0pt}
In this paper, we introduce HTCL, an innovative hierarchical temporal in-context learning paradigm for semantic scene completion (SSC). 
To highlight the most relevant patterns, we introduce the pattern affinity to measure the contextual correspondence between the current and historical frames. 
Subsequently, to dynamically compensate for incomplete observations, we propose to adaptively refine the feature sampling locations based on the initially high-affinity locations and their neighboring relevant regions. 
Our method outperforms state-of-the-art camera-based methods and even surpasses LiDAR-based methods for semantic scene completion. We hope HTCL could inspire further research in camera-based temporal modeling for SSC and its applications in 3D visual perception.

\section*{Acknowledgments}
This work was supported in part by NSFC 62302246 and ZJNSFC under Grant LQ23F010008, and supported by High Performance Computing Center at Eastern Institute of Technology, Ningbo, and Ningbo Institute of Digital Twin.

\newpage
\appendix
\vspace{50pt}
\section*{\Large \centering Supplementary Material for HTCL}
 \vspace{30pt}

\section{Additional Results}

\subsection{Qualitative Results}
\begin{figure*}[!ht]
   \vspace{-0pt}
	\begin{center}
		\includegraphics[width=12.0
  cm]{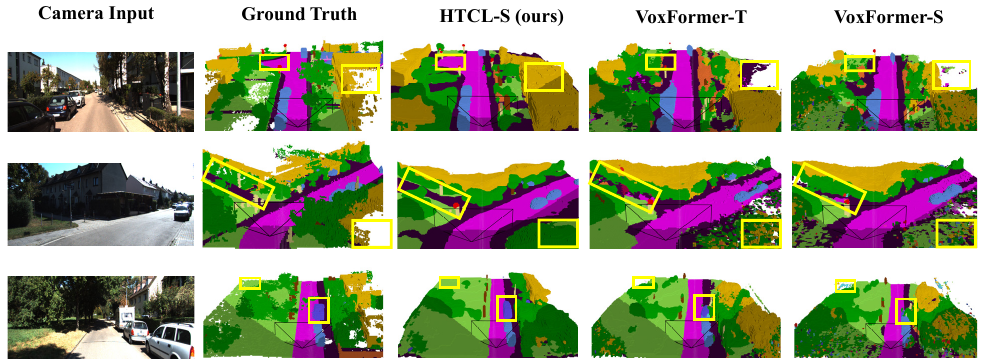}   
        \vspace{0pt}
		\begin{tabular}{cccccc}
			\multicolumn{6}{c}{
				\scriptsize
				\textcolor{bicycle}{$\blacksquare$}bicycle~
				\textcolor{car}{$\blacksquare$}car~
				\textcolor{motorcycle}{$\blacksquare$}motorcycle~
				\textcolor{truck}{$\blacksquare$}truck~
				\textcolor{other-vehicle}{$\blacksquare$}other-veh.~
				\textcolor{person}{$\blacksquare$}person~
				\textcolor{bicyclist}{$\blacksquare$}bicyclist~
				\textcolor{motorcyclist}{$\blacksquare$}motorcyclist~
				\textcolor{road}{$\blacksquare$}road~
				}
    \\
			\multicolumn{6}{c}{
				\scriptsize
                    \textcolor{parking}{$\blacksquare$}parking \textcolor{sidewalk}{$\blacksquare$}sidewalk \textcolor{other-ground}{$\blacksquare$}other.grd \textcolor{building}{$\blacksquare$}building \textcolor{fence} {$\blacksquare$}fence \textcolor{vegetation}{$\blacksquare$}vegetation \textcolor{trunk}{$\blacksquare$}trunk \textcolor{terrain}{$\blacksquare$}terrain \textcolor{pole}{$\blacksquare$}pole \textcolor{traffic-sign}{$\blacksquare$}traf.sign	
			}
		\end{tabular}	
 	\end{center}
        \vspace{-0pt}
	\caption{ \textbf{Qualitative results} on the SemanticKITTI validation set.} 
	\label{fig_q1}
 \vspace{-0pt}
\end{figure*}

\begin{figure}[!ht]
   \vspace{-0pt}
	\begin{center}
		\includegraphics[width=12cm]{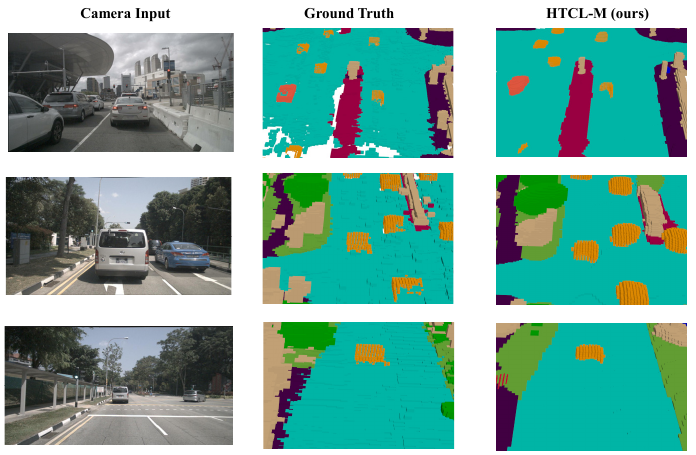}   
        \vspace{0pt}
		\begin{tabular}{cccccccc}
			\multicolumn{8}{c}{
				\scriptsize
				\textcolor{bicycle1}{$\blacksquare$}bicycle \textcolor{bus1}{$\blacksquare$}bus \textcolor{car1}{$\blacksquare$}car \textcolor{const. veh.1}{$\blacksquare$}const.veh. \textcolor{motorcycle1}{$\blacksquare$}motorcycle \textcolor{pedestrian1}{$\blacksquare$}pedestrian \textcolor{truck1}{$\blacksquare$}truck \textcolor{sidewalk1}{$\blacksquare$}sidewalk~
        
				}
    \\
			\multicolumn{7}{c}{
				\scriptsize

                \textcolor{traffic cone1}{$\blacksquare$}traffic cone \textcolor{trailer1}{$\blacksquare$}trailer \textcolor{drive. suf.1}{$\blacksquare$}drive.suf. \textcolor{other flat1}{$\blacksquare$}other flat \textcolor{terrain1}{$\blacksquare$}terrain \textcolor{manmade1}{$\blacksquare$}manmade \textcolor{vegetation1}{$\blacksquare$}vegetation~
			}
		\end{tabular}	
 	\end{center}
        \vspace{-0pt}
	\caption{ \textbf{Qualitative results} on the OpenOccupancy validation set.} 
	\label{fig_open1}
 \vspace{-0pt}
\end{figure}

We report additional qualitative results in Figure~\ref{fig_q1} and Figure~\ref{fig_open1}.
As illustrated in Figure~\ref{fig_q1}, we provide more qualitative comparison results between our proposed method and VoxFormer on the SemanticKITTI~\cite{behley2019semantickitti} validation set. 
Compared with VoxFormer, our method can predict more accurate scene layouts (e.g., crossroads in the first and second rows) and moving objects (e.g., trucks in the third row).
Moreover, our method hallucinates more complete and proper out-FOV (field of view) scenes(e.g., shadow areas in the second row). 
As illustrated in Figure~\ref{fig_open1}, we also report more visualization results on the OpenOccupancy~\cite{wang2023openoccupancy} validation set. Compared to the ground truth with sparse annotations, our proposed method can generate more fine-grained realistic predictions (e.g., dense road predictions in the first and second rows).

 \subsection{Quantitative Results}
As shown in Table~\ref{tabq2}, We conduct additional quantitative experiments on the SemanticKITTI validation dataset with other camera-based SSC methods~\cite{li2023voxformer,cao2022monoscene}.
Compared to other baselines, our method achieves significant improvements in mIoU, demonstrating our method's effectiveness for semantic scene completion. Specifically, our method shows obvious superiority in capturing better moving objects (e.g., cars, bicycles, trucks) and scene layouts (e.g., roads, sidewalks).

\begin{table*}[!ht]
\vspace{-0pt}
\begin{center}
\scriptsize
\caption{\textbf{Quantitative results} on the SemanticKITTI validation set with the state-of-the-art camera-based SSC methods. 
The {``S-T'', ``S''} and {``M''} denote temporal stereo images, single-frame stereo images, and single-frame monocular images, respectively.  
}
\vspace{-0pt}
\renewcommand\tabcolsep{8.1pt}
\resizebox{0.98\textwidth}{!}{%
\begin{tabular}{l|c|c|c|c}
\toprule
{Methods}   & HTCL-S (ours)  & VoxFormer-T  & VoxFormer-S   & MonoScene  \\ \midrule

{Input}     & {S-T}  &{S-T} &{S}  &{M}   \\ \midrule
\textbf{IoU }  & \textbf{45.51}   & 44.15 &44.02 &37.12 \\ \midrule
\textbf{mIoU }      & \textbf{17.13}  &13.35 &12.35 &11.50   \\ \midrule

\crule[carcolor]{0.13cm}{0.13cm} {car}  & \textbf{34.30}   & 26.54 & 25.79 & 23.55 \\

\crule[bicyclecolor]{0.13cm}{0.13cm} {bicycle}  & \textbf{3.99}  & 1.28 & 0.59 & 0.20 \\

\crule[motorcyclecolor]{0.13cm}{0.13cm} {motorcycle}    & \textbf{2.80} & 0.56 & 0.51 &0.77  \\

\crule[truckcolor]{0.13cm}{0.13cm} {truck} & \textbf{20.72}& 8.10 & 7.26 & 7.83     \\

\crule[othervehiclecolor]{0.13cm}{0.13cm} {other-veh.} & \textbf{11.99}  &7.81 &3.77 &3.59  \\

\crule[personcolor]{0.13cm}{0.13cm} {person}  & \textbf{2.56} &1.93 &1.78 &1.79  \\

\crule[bicyclistcolor]{0.13cm}{0.13cm} {bicyclist}  & {2.30} &1.97 &\textbf{3.32} &1.03   \\

\crule[motorcyclistcolor]{0.13cm}{0.13cm} {motorcyclist} & {0.00} &{0.00}  &0.00 &{0.00}  \\

\crule[roadcolor]{0.13cm}{0.13cm} {road}  & \textbf{63.70}  &53.57 &54.76 &57.47    \\

\crule[parkingcolor]{0.13cm}{0.13cm} {parking}  & \textbf{23.27}  &{19.69} &15.50 &15.72  \\

\crule[sidewalkcolor]{0.13cm}{0.13cm} {sidewalk}    & \textbf{32.48}  &26.52 &26.35 &27.05    \\

\crule[othergroundcolor]{0.13cm}{0.13cm} {other.grd} & {0.14}&0.42 &0.70 &\textbf{0.87}    \\

\crule[buildingcolor]{0.13cm}{0.13cm} {building}   &\textbf{24.13} & {19.54}  &17.65 &14.24   \\

\crule[fencecolor]{0.13cm}{0.13cm} {fence}  & \textbf{11.22}  &{7.31} &7.64 &6.39  \\

\crule[vegetationcolor]{0.13cm}{0.13cm} {vegetation}   & \textbf{26.96}  &26.10 &24.39 &18.12  \\

\crule[trunkcolor]{0.13cm}{0.13cm} {trunk}  & \textbf{8.79}  &6.10 &5.08 &2.57  \\

\crule[terraincolor]{0.13cm}{0.13cm} {terrain}   & \textbf{37.73}  &33.06 &29.96 &30.76   \\

\crule[polecolor]{0.13cm}{0.13cm} {pole}  & \textbf{11.49}  &9.15 &7.11 &4.11  \\

\crule[trafficsigncolor]{0.13cm}{0.13cm} {traf.sign}   & \textbf{6.95}   &4.94 &4.18 &2.48 \\  \bottomrule

\end{tabular}
}
\label{tabq2}        
\vspace{-0pt}
\end{center}
\end{table*}

\section{More Visualization on Cross-frame Pattern Affinity}
\begin{figure}[t]
\vspace{-0pt}
\centering
\includegraphics[width=12.0cm]{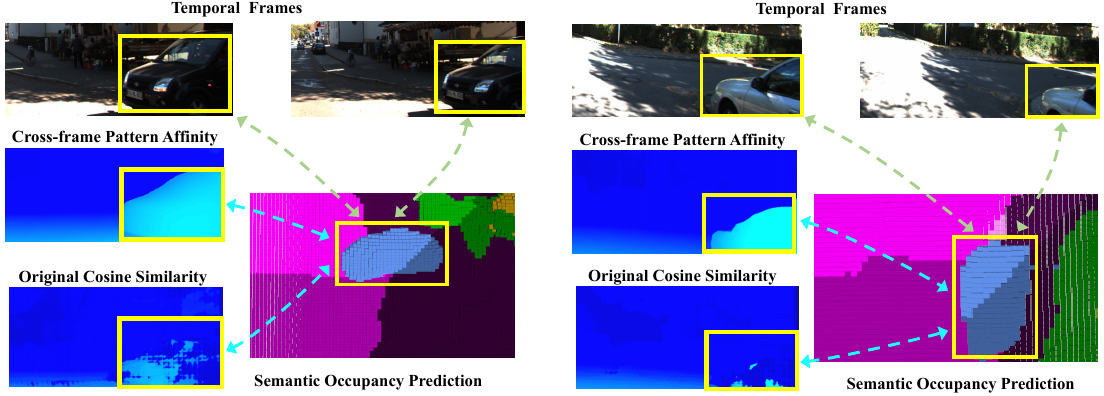}  
\vspace{-0pt}
\caption{Visualization of the heat maps from our proposed Cross-frame Pattern Affinity (CPA) and the original cosine similarity. 
} 
\label{fig_aff2}
\vspace{-0pt}
\end{figure}

We provide more visualization results in Figure~\ref{fig_aff2}. Compared with the original cosine similarity, our proposed Cross-frame Pattern Affinity (CPA) effectively illustrates the contextual correspondence within the temporal content.

\section{Extensive Experiments on BEV Detection}

\begin{figure}[!ht]
   \vspace{-0pt}
	\begin{center}
        \vspace{-0pt}
        \includegraphics[width=0.8\linewidth]{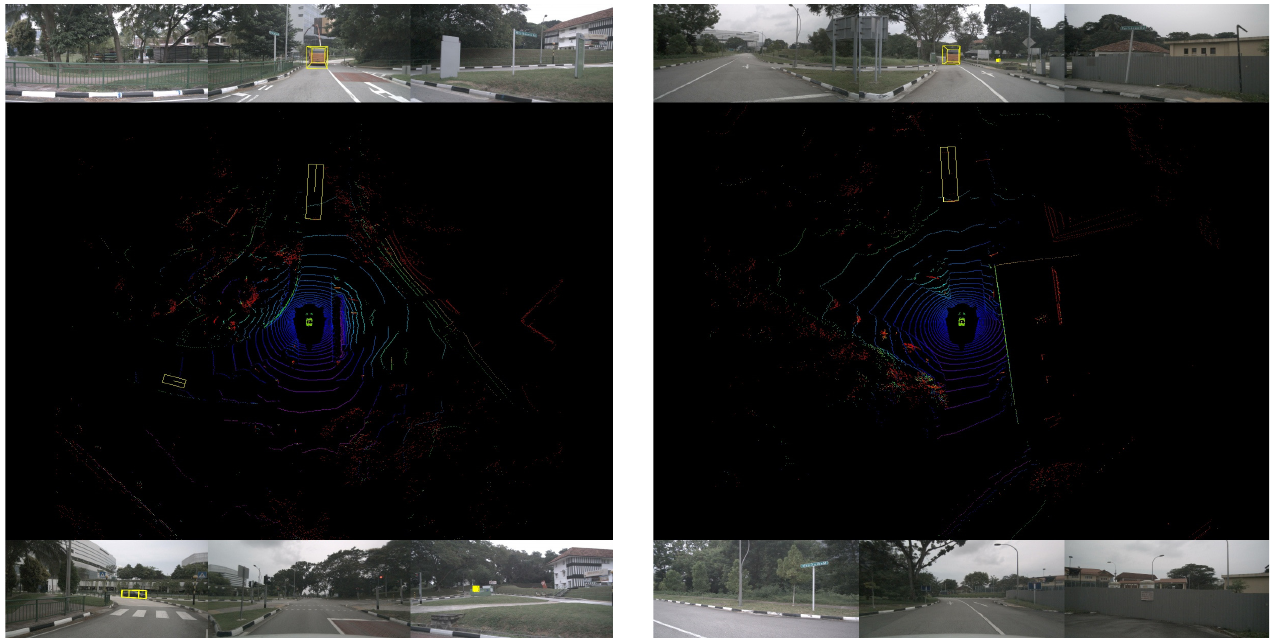}
        \end{center}
        \vspace{-0pt}
	\caption{\textbf{Visualization results} of BEV detection on the nuScenes validation set. } 
	\label{fgvisdet2}
 \vspace{-0pt}
\end{figure}

To further demonstrate the potential of our method, we present preliminary experimental evaluations on the Bird-Eye-View (BEV) detection~\cite{zhang2019exploiting,li2021enforcing,kopf2021robust,lin2022sparse4d,liu2022petr} in the nuScenes~\cite{caesar2020nuscenes} validation dataset. Specifically, we utilize \textit{BEVDet}~\cite{huang2021bevdet} as the baseline configuration and substitute the original model of \textit{BEVDet} with our proposed \textit{HTCL-M}, while maintaining the same detection head. 
The evaluation results are reported in Table~\ref{tab_bev} and visually depicted in Figure~\ref{fgvisdet2}. 
As shown in Table~\ref{tab_bev}, our HTCL-M outperforms BEVDet4D-Base with a relative improvement of 20.19\% mAP and 6.34\% NDS, respectively.
These results illustrate the effectiveness of our proposed methodology, indicating its potential applicability to a broader spectrum of downstream tasks.

\begin{table}[!ht]\centering
\small
\renewcommand\tabcolsep{16.5pt}
\caption{ \textbf{Quantitative results} of BEV Detection on the nuScenes validation set. We conduct preliminary experiments by employing the detection head. }
\begin{tabular}{ l |ccc}
\toprule
 \textbf{Methods}& \textbf{Resolution}& \textbf{mAP} $\uparrow$  & \textbf{NDS} $\uparrow$   \\  \midrule

BEVDet-Base  & 1600$\times$ 640 & 0.397 & 0.477 \\

BEVDet4D-Base  & 1600$\times$ 640 & 0.426 & 0.552 \\

PETR-R101  & 1408$\times$ 512 & 0.357 & 0.421\\

BEVDepth-R101  & 512$\times$ 1408 & 0.412 & 0.535 \\

\rowcolor{gray!10} HTCL-M (ours) & 1600$\times$ 640  &\textbf{0.512} &\textbf{0.587} \\  
  \bottomrule
\end{tabular}
\vspace{-0pt}
\label{tab_bev}
\end{table}

\section{Additional Ablation Studies}
We conduct additional ablation studies on the Multi-group Context Generation and the Multi-level Deformable Block, as presented in Table~\ref{tab_ab}.
As introduced in the main paper, we employ multiple groups of contextual features to facilitate diverse independent similarity learning.
The results in Table~\ref{tab_ab} demonstrate that leveraging 3 contextual groups yields a significant performance improvement, while employing more groups (5 groups) leads to a relatively slight improvement.
Similarly, the enhancement of utilizing more feature levels (5 levels) in the Multi-level Deformable Block is also relatively minor.
Therefore, considering the time consumption and parameter efficiency, we adopt 3 contextual groups in the Multi-group Context Generation and 3 feature levels in the Multi-level Deformable Block as the default settings.

\begin{table*}[!ht]\centering
\vspace{-0pt}
\renewcommand\tabcolsep{1.4pt}
\scriptsize
\caption{ {Ablation studies} on the Multi-group Context Generation and the Multi-level Deformable Block. }
\resizebox{0.98\textwidth}{!}{
\begin{tabular}{ccc|ccc|cc} \toprule
\multicolumn{3}{c|}{\textbf{Multi-group Context Generation}} & \multicolumn{3}{c|}{\textbf{Multi-level Deformable Block}} & \multirow{2}{*}{\textbf{mIoU (\%)}} & \multirow{2}{*}{\textbf{Time (s)}} \\

\qquad 1 Group  &\qquad 3 Groups &\qquad 5 Groups & \qquad1 Level \qquad  &3 Levels \qquad  &5 Levels \qquad & & \\  \midrule 
\qquad \checkmark \qquad  & & & & \qquad \checkmark & & 15.26 &  0.283      \\ 
\qquad &\qquad &\qquad  \checkmark  & & \qquad \checkmark &  & 17.21 &  0.312 \\   

 \midrule

& \qquad \checkmark &  & \qquad \checkmark & \qquad & & 16.51 & 0.286 \\ 

& \qquad \checkmark &  & \qquad  & \qquad  & \qquad \checkmark & 17.18  & 0.309  \\  \midrule
\rowcolor{gray!10} \qquad  & \qquad \checkmark &\qquad  & \qquad  &\qquad \checkmark & \qquad  &{17.13} &  {0.297} \\  \bottomrule
\end{tabular}
}
\vspace{-0pt}
\label{tab_ab}
\end{table*}

\section{Loss Function}

\noindent \textbf{PoseNet implementation}
We implement the PoseNet following previous video depth estimation methods~\cite{guizilini20203d,watson2021temporal}. 
To reduce the learning burden, we pre-train the PoseNet and freeze it for temporal semantic occupancy learning.
The PoseNet is trained without ground truth in a self-supervised manner.
At the pre-training stage, the training objective of the PoseNet is:

\begin{equation}
\mathcal{L}_{pose} = \min _n \mathrm{PE} \left(I_t, I_{t+n \rightarrow t}\right) + \lambda*\mathcal{L}_{smooth}.
\end{equation}
where $\mathrm{PE}$ is a combination of SSIM and L1 losses between reference image $I_t$ and source image $I_{t+n}$. ${L}_{smooth}$ is the smoothness loss for pixel-level regularization from~\cite{guizilini20203d,watson2021temporal}. $\lambda$ is the balance coefficient.
We will add the details in Section 7 (Network Training) of the supplementary material.

\noindent \textbf{Network Training.} 
We follow the basic learning objective of MonoScene~\cite{cao2022monoscene} for semantic scene completion. 
Standard semantic loss $\mathcal{L}_{\text{sem}}$ and geometry loss $\mathcal{L}_{\text{geo}}$ are leveraged for semantic and geometry supervision, while an extra class weighting loss $\mathcal{L}_{ce}$ is also added.
To further enforce the ensembled volume, we adopt a binary cross entropy loss $\mathcal{L}_{depth}$ to encourage the sparse depth distribution. The overall learning objective of this framework is formulated as:
\begin{equation}
   { \mathcal{L} = \mathcal{L}_{depth} + \mathcal{\lambda}_{ce} \mathcal{L}_{ce}. }
\end{equation}
where several $\lambda$s are balancing coefficients.

\section{Limitation and Potential Negative Impact}
The running speed of our model could be further enhanced as more lightweight networks are more practical for real-world applications. We leave this to our future work. 
While the promising semantic scene completion results of the proposed method could promote the development of autonomous driving, the legal challenges, as well as the privacy and data security risks of autonomous driving remain subjects of debate.

\bibliographystyle{splncs04}
\bibliography{main}
\end{document}